\documentclass[twoside, twocolumn]{article}
\pdfoutput=1

\newlength\titlebox 
\renewenvironment{abstract}
   {%
\centerline{\large\bf Abstract}
    \vspace{-0.12in}\begin{quote}}
   {\par\end{quote}\vskip 0.12in}

\usepackage{hyperref}

\usepackage[normalem]{ulem}

\usepackage{authblk}
\usepackage{graphicx} 
\usepackage{wrapfig}
\usepackage{subfig}
\usepackage{floatrow} 

\usepackage{array}

\usepackage{xspace}

\usepackage{enumitem}
\usepackage{algorithm}
\usepackage{algorithmic}

\usepackage{amsmath}
\usepackage[T1]{fontenc}
\usepackage[utf8x]{inputenc}
\usepackage{amsfonts} 
\usepackage{amsthm} 

\usepackage[svgnames]{xcolor}

\newcommand{\rcolbm}[1]{$\textcolor{red}{\boldsymbol{#1}}$}

\newlength{\minipagewidth}
\newlength{\minipagewidthx}
\newcommand{\bookboxx}[1]{\small
\par\medskip\noindent
\framebox[\columnwidth]{
\begin{minipage}{0.98\columnwidth} {\par\smallskip#1\par\smallskip} \end{minipage} } \par\medskip }

\def\:#1{\mathbf{#1}}
\newcommand{\hfssparse}{\textsc{Sparse-HFS}\xspace}
\newcommand{\hfsstable}{\textsc{Stable-HFS}\xspace}
\newcommand{\hfs}{\textsc{HFS}\xspace}
\newcommand{\simplehfs}{\textsc{Simple-HFS}\xspace}
\newcommand{\subs}{\textsc{SubSampling}\xspace}
\newcommand{\fergus}{\textsc{EigFun}\xspace}
\newcommand{\onenn}{\textsc{1-NN}\xspace}

\newcommand{\calS}{\mathcal{S}}
\newcommand{\X}{\mathcal{X}}
\newcommand{\calP}{\mathcal{P}}

\renewcommand{\Re}{\mathbb{R}}

\newcommand{\wt}[1]{\widetilde{#1}}
\newcommand{\wh}[1]{\widehat{#1}}
\newcommand{\ol}[1]{\overline{#1}}
\newcommand{\transp}{\mathsf{T}}
\DeclareMathOperator*{\argmin}{arg\,min}

\DeclareMathOperator*{\Ker}{Ker}

\newcommand{\normsmall}[1]{\Vert #1 \Vert}

\newcommand{\Real}{\mathbb{R}}

\newcommand{\Gg}{\mathcal{G}}
\newcommand{\Hg}{\mathcal{H}}

\newcommand{\edgeset}{\mathcal{E}}

\newcommand{\funcspace}{\mathcal{F}}

\newcommand{\trainingset}{\mathcal{S}}
\newcommand{\testset}{\mathcal{T}}

\newcommand{\vareps}{\varepsilon}
\newcommand{\bigotime}{\mathcal{O}}
\DeclareMathOperator*{\polylog}{polylog}

\newtheorem{theorem}{Theorem}
\newtheorem{lemma}{Lemma}
\newtheorem{definition}{Definition}

\title{\textbf{Incremental Spectral Sparsification for Large-Scale Graph-Based Semi-Supervised Learning}}

\author[1]{Daniele Calandriello\thanks{daniele.calandriello@inria.fr}}
\author[1]{Alessandro Lazaric\thanks{alessandro.lazaric@inria.fr}}
\author[1]{Michal Valko\thanks{michal.valko@inria.fr}}
\affil{SequeL team, INRIA Lille - Nord Europe, France}
\author[2]{Ioannis Koutis\thanks{ioannis.koutis@upr.edu}}
\affil[2]{Computer Science Department,
University of Puerto Rico - Rio Piedras}

\date{}

\begin{document}
\maketitle

\begin{abstract}
While the \textit{harmonic function solution} performs well in many semi-supervised learning (SSL) tasks, it is known to scale poorly with the number of samples.
Recent successful and scalable methods, such as the eigenfunction method~\cite{fergus2009semi-supervised} focus on efficiently approximating the whole spectrum of the graph Laplacian constructed from the data. This is in contrast to various subsampling and quantization methods proposed in the past, which may fail in preserving the graph spectra. However, the impact of the approximation of the spectrum on the final generalization error is either unknown~\cite{fergus2009semi-supervised}, or requires strong assumptions on the data~\cite{yang2012simple}.  In this paper, we introduce \hfssparse, an efficient edge-sparsification algorithm for SSL. By constructing an edge-sparse and spectrally similar graph, we are able to leverage the approximation guarantees of spectral sparsification methods to bound the generalization error of \hfssparse. As a result, we obtain a theoretically-grounded approximation scheme for graph-based SSL that also empirically matches the performance of known large-scale methods.
\end{abstract}

\vspace{-0.05in}
\section{Introduction}
\vspace{-0.05in}
In many classification and regression tasks, obtaining labels for large
datasets is expensive. When the number of labeled samples is too small,
traditional supervised learning algorithms fail in learning accurate
predictors. \emph{Semi-supervised learning} (SSL)
\cite{chapelle2010semi-supervised,zhu2008semi-supervised} effectively
deals with this problem by integrating the labeled samples with
an additional set of unlabeled samples, which are
abundant and readily available in many applications (e.g., set of images collected on web
sites~\cite{fergus2009semi-supervised}). The intuition behind SSL is that
unlabeled data may reveal the underlying structure of the problem (e.g., a
manifold) that could be exploited to compensate for the limited number of
labels and improve the prediction accuracy.
Among different SSL settings, in this paper we focus on the case where data
are embedded in a \emph{graph}.
The graph is expected to effectively represent the geometry of data and
graph-based SSL~\cite{zhu2003semi-supervised,belkin2006manifold,subramanya2014graph} methods
leverage the intuition that nodes that are similar according to the graph are
more likely to be labeled similarly.
A popular approach is the \emph{harmonic function solution} (\hfs)
\cite{zhu2003semi-supervised,belkin2004regularization,fergus2009semi-supervised},
whose objective is to find a solution where each node's value is the weighted
average of its neighbors. Computing the \hfs solution requires solving a
Laplacian regularized least-squares problem.
While the resulting solution is
both empirically effective~\cite{fergus2009semi-supervised} and enjoys strong
performance guarantees~\cite{belkin2006manifold,cortes2008stability},
solving \emph{exactly} the least-squares problem on a graph with $n$ nodes
amounts to $\bigotime(n^3)$ time and $\bigotime(n^2)$ space complexity. Using a general iterative
solver on a graph with $m$ edges to obtain a comparable solution requires
only $\bigotime(mn)$ time and $\bigotime(m)$ space,
but this is still practically unfeasible in many applications of
interest. In fact, many graphs have naturally a large number of edges, so that
even if the $n$ nodes could fit in memory, storing $m$ edges largely exceeds
the memory capacity. For instance, Facebook's graph of
relationships~\cite{ching2015one} counts about $n=1.39e9$ users connected by a
trillion ($m=1e12$) edges. While $n$ is still
in the order of the memory capacity, the edges cannot be stored in a single
computer.
A similar issue is faced when
the graph is built starting from a dataset, for instance using
a $k$-nn graph. In this case $m=kn$ edges are created, and sometimes a large
$k$ is necessary to obtain good performance~\cite{saluja2014graph},
or artificially adding neighbours can improve the stability of the
method \cite{Gleich2015robustifying}.
In such problems, a direct application of \hfs is not possible and thus some
form of approximation or graph sketching is required. A straightforward
approach is to distribute the graph over multiple machines in a cluster and
resort to an iterative solver for the solution of the least-squares
problem~\cite{ching2015one}. Distributed algorithms require infrastructure and
careful engineering in order to deal with communication issues
\cite{cai2014comparison}. But even assuming that these problems are
satisfactorily dealt with, all known iterative solvers that
have provably fast convergence
do not have known distributed implementations, as they assume
random, constant time access to all the edges in the graph. Thus one would have to
resort to distributed implementations of simpler but much slower
methods, in effect trading-off space for a significant reduction in overall
efficiency.

More principled
methods try to address the memory bottleneck by directly manipulating the
structure of the graph to reduce its size.
These include \emph{subsampling} the
nodes of the original graph, \emph{quantization}, approaches related to
\emph{manifold learning}, and various \emph{approximation} strategies.
The most straightforward way to reduce the complexity in graph-based method is
to subsample the nodes to create a smaller, \emph{backbone graph} of
representative vertices, or \emph{landmarks}~\cite{talwalkar2008large-scale}.
Nystr\"{o}m sampling methods~\cite{kumar2012sampling} randomly select $s$
nodes from the original graph and compute $q$ eigenvectors of the smaller
graph, which can be later used to solve the HFS regularized problem. It can be
shown~\cite{kumar2012sampling} that the reconstructed Laplacian is accurate in
$\ell_2$-norm and thus only its largest eigenvalue is preserved.
Unfortunately, the \hfs solution does not depend only on the largest
eigenvalues, both because the largest eigenvectors are the ones most penalized
by \hfs's regularizer (\cite{belkin2006manifold}) and because theoretical
analysis shows that preserving the smallest eigenvalue is important for generalization
bounds~(\cite{belkin2004regularization}).
As a result, subsampling methods can completely fail when the sampled nodes
compromise the spectral structure of the
graph~\cite{fergus2009semi-supervised}. Although alternative techniques have
been developed over years (see e.g.,
\cite{jebara2009graph,yu2005blockwise,zhu2005harmonic,garcke2005semi-supervised,tsang2006large-scale,liu2010large}),
this drawback is common to all backbone graph methods.
Motivated by this observation, other approaches focus on computing a more
accurate approximation of the spectrum of the Laplacian. Fergus \textit{et
    al.}~\cite{fergus2009semi-supervised} build on the observation that when
the number of unlabeled samples$n$ tends to infinity, the eigenvectors of the
Laplacian tend to the eigenfunctions of the sampling distribution $\calP$.
Thus instead of approximating eigenvectors in $\Re^n$, they first compute
empirical
eigenfunctions of the estimated sampling distribution (defined on the $d$-dimensional
feature space) obtained by assuming that~$\calP$ is factorized
and by using a histogram estimation over $b$ bins over each dimension
separately. While the method scales to the order of million nodes, it still
requires $d$ and $b$ to be small to be efficient. Furthermore, no theoretical analysis is
available, and the method  may return poor approximations whenever the
sampling distribution is not factorized.
Motivated by the empirical success of~\cite{fergus2009semi-supervised},
Ji \textit{et al.}~\cite{yang2012simple} proposed a similar algorithm,
\simplehfs, for which they provide theoretical guarantees.
However, in order to prove bounds on the generalization error, they need to
assume several strong and hard to verify assumptions, such as a sufficiently
large eigengap. On the contrary, the guarantees for our method work for any graph.

\paragraph{Our contribution}
In this paper, we focus on reducing the space complexity of graph-based SSL while matching the smallest possible computational complexity of $\Omega(m)$ up to logarithmic factors\footnote{While the computational complexity of exact \hfs is $\bigotime(mn)$, many approximated methods can significantly reduce it. Nonetheless, any method that requires reading all the edges once has at least $\Omega(m)$ time complexity.} and providing strong guarantees about the quality of the solution.
In particular, we introduce a novel approach which employs efficient
spectral graph sparsification techniques~\cite{kelner_spectral_2013} to
incrementally process the original graph. This method, coupled with dedicated
solvers for symmetric diagonally dominant (SDD)
systems~\cite{koutis2011a-nearly-m}, allows to find an approximate \hfs
solution without storing the whole graph in memory and to control the
computational complexity as $n$ grows. In fact, we show that
our proposed method, called \hfssparse, requires only fixed $\bigotime(n \log^2(n))$
space to run, and allows to compute solutions to large \hfs problems in memory.
For example, in the experimental section we show that the sparsifier can
achieve an accuracy comparable to the full graph, using one order
of magnitude less edges. With a careful choice of the frequency of
resparsification, the proposed method does not
increase significantly the running time. Given a minimum amortized cost of
$\Omega(1)$ per edge, necessary to examine each edge at least once, our algorithm only increases
this cost to $\bigotime(\log^3(n))$.
Furthermore, using the approximation properties of spectral sparsifiers and
results from algorithmic stability theory~\cite{bousquet_stability_2002,
    cortes2008stability} we provide theoretical guarantees for the
generalization error for \hfssparse, showing that the performance is
asymptotically the same as the exact solution of \hfs. Finally, we report
empirical results on both synthetic and real data showing that \hfssparse is
competitive with subsampling and the \fergus method
in~\cite{fergus2009semi-supervised}.

\vspace{-0.05in}
\section{Graph-Based Semi-Supervised Learning}
\vspace{-0.05in}

\textbf{Notation.}
We denote with lowercase letter $a$ a scalar, with bold lowercase letter $\:a$
a vector and with uppercase letter $A$ a matrix.
We consider the problem of regression
in the semi-supervised setting, where a large set of $n$ points
$\X = (\:x_1,\ldots,\:x_n) \subset \Re^d$ is drawn from a distribution $\calP$
and labels $\{y_i\}_{i=1}^l$ are provided only for a small (random) subset
$\calS \subset \X$ of $l$ points. Graph-based SSL builds on the observation
that $\calP$ is often far from being uniform and it may display a specific
structure that could be exploited to ``propagate'' the labels to similar
unlabeled points. Building on this intuition, graph-based SSL algorithms
consider the case when the points in $\X$ are embedded into an undirected
weighted graph $\Gg = (\X,\edgeset)$ with $|\edgeset| = m$ edges. Associated
with each edge $e_{i,j}\in \edgeset$ there is a weight~$a_{e_{i,j}}$ measuring
the ``distance'' between $\:x_i$ and$\:x_j$\footnote{Notice that $\Gg$ can be either constructed from the data (e.g.,
    building a $k$-nn graph using the exponential distance
    $a_{e_{i,j}} = \exp(- ||x_i - x_j||_2 / \sigma^2)$) or it can be provided
    directly as input (e.g., in social networks).}.
A graph-based SSL algorithm receives as input $\Gg$ and the labels of the
nodes in $\calS$ and it returns a function $\:f : \X \rightarrow \Re$ that
predicts the label for all nodes in~$\X$. The objective is to minimize the
prediction error over the set~$\mathcal{T}$ of $u = n-l$ unlabeled nodes. In
the following we denote by $\:y\in\Re^n$ the full vector of labels.

\textbf{\hfsstable.}
\hfs directly exploits the structure embedded in $\Gg$ to learn functions that
are smooth over the graph, thus predicting similar labels for similar nodes.
Given the weighted adjacency matrix $A_{\Gg}$ and the degree matrix $D_{\Gg}$,
the Laplacian of $\Gg$ is defined as $L_\Gg = D_\Gg - A_\Gg$. The Laplacian
$L_{\Gg}$ is semi-definite positive (SDP) with $\Ker(L_\Gg)=\:1$.
Furthermore, we assume that $\Gg$ is connected and thus has only one
eigenvalue at $0$. Let $L_{\Gg}^+$ be the pseudoinverse of $L_{\Gg}$, and
$L_{\Gg}^{-1/2} = (L_{\Gg}^+)^{1/2}$. The \hfs
method~\cite{zhu2003semi-supervised} can be formulated as
the Laplacian-regularized least-squares problem
\begin{align}\label{eq:hfs.original}
    \wh{\:f} &= \argmin_{\:f \in \Re^n}  \tfrac{1}{l} (\:f - \:y)^\transp I_{\trainingset} (\:f-\:y) + \gamma \:f^\transp L_\Gg \:f,
\end{align}
where $I_{\trainingset}\in\Re^{n\times n}$ is the identity matrix with zeros
corresponding to the nodes not in $\trainingset$ and $\gamma$ is a
regularization parameter. The solution can be computed in closed form as
$\wh{\:f} = (\gamma l L_\Gg + I_\trainingset)^+ \:y_\calS$, where
$\:y_\calS = I_\calS \:y \in \Re^n$.
The singularity of the Laplacian may lead to unstable behavior with
drastically different results for small perturbations to the dataset. For this
reason, we focus on the \hfsstable algorithm proposed
in~\cite{belkin2004regularization} where an additional regularization term is
introduced to restrict the space of admissible hypotheses to the space
$\funcspace = \left\{\:f : \langle\:f,\:1\rangle = 0\right\}$ of functions
orthogonal to null space of $L_\Gg$ (i.e., centered functions). This
restriction can be easily enforced by introducing an additional regularization
term $\frac{\mu}{l} \:f^\top \:1$ in Eq.~\ref{eq:hfs.original}. As shown
in~\cite{belkin2004regularization}, in order to guarantee that the resulting
$\wh{\:f}$ actually belongs to $\funcspace$, it is sufficient to set the
regularization parameter to
$\mu = ((\gamma l L_\Gg + I_\trainingset)^+\:y_S)^\transp \:1 / ((\gamma l L_\Gg + I_\trainingset)^+ \:1)^\transp \:1$,
and compute the solution as
$\wh{\:f} = (\gamma l L_\Gg + I_\trainingset)^+ (\:y_\calS - \mu \:1)$.
Furthermore, it can be shown that if we center the vector of labels
$\wt{\:y}_\calS = \:y_\calS - \ol{\:y}_\calS$, with
$\ol{\:y} = \frac{1}{l}\:y_\calS^\transp \:1$, then the solution of \hfsstable
can be rewritten as
$\wh{\:f} = (\gamma l L_\Gg + I_\trainingset)^+ (\wt{\:y}_\calS - \mu \:1) = \big(P_\funcspace (\gamma l L_\Gg + I_\trainingset)\big)^+ \wt{\:y}_\calS$,
where $P_\funcspace = L_\Gg L_\Gg^+$ is the projection matrix on the $n-1$
dimensional space $\funcspace$. Indeed, since the Laplacian of any graph $\Gg$
has a null space equal to the one vector $\:1$, then $P_\funcspace$ is
invariant w.r.t.\@ the specific graph $\Gg$ used to defined it. While \hfsstable
is more stable and thus more suited for theoretical analysis, its
time and space requirements remain $\bigotime(mn)$ and $\bigotime(m)$,
and cannot be applied to graph with a large number of edges.

\vspace{-0.05in}
\section{Spectral Sparsification for Graph-Based SSL}
\vspace{-0.05in}

\begin{figure}
\vspace{-0.2in}
\begin{minipage}[t]{1.0\linewidth}
\bookboxx{
\begin{algorithmic}
\begin{small}
    \INPUT Graph $\Gg = (\X, \edgeset)$, labels $\:y_\calS$, accuracy $\vareps$
    \OUTPUT Solution $\wh{\:f}$, sparsified graph $\Hg$
    \STATE Let $\alpha = 1/(1-\vareps)$ and $N = \alpha^2 n \log^2(n)/\vareps^2$
    \STATE Partition $\edgeset$ in $\tau = \lceil m / N\rceil$ blocks $\Delta_1,\ldots,\Delta_\tau$
    \STATE Initialize $\Hg = \emptyset$
    \FOR{$t=1,\ldots,\tau$}
        \STATE Load $\Delta_t$ in memory
        \STATE Compute $\Hg_t = \textsc{sparsify}(\Hg_{t-1}, \Delta_t, N, \alpha)$
    \ENDFOR
    \STATE Center the labels $\wt{\:y}_\calS$
    \STATE Compute $\wt{\:f}$ with \hfsstable with $\wt{\:y}_\calS$ using a suitable SDD solver
\end{small}
\end{algorithmic}
}
\vspace{-0.1in}
\caption{\small\hfssparse\vspace{-0.1in}}\label{alg:sparse_ssl}
\end{minipage}
\end{figure}

In this section we introduce a novel variant of \hfs, called \textit{\hfssparse}, where spectral graph sparsification techniques are integrated into \hfsstable, drastically reducing the time and memory requirements without compromising the resulting accuracy.

\textbf{Spectral sparsification.} 
A graph sparsifier receives as input a graph
$\Gg$ and it returns a graph $\Hg$ on the same set of nodes $\X$ but with much fewer edges. Among different techniques~\cite{Batson:2013:SSG:2492007.2492029},
spectral sparsification methods provide the stronger guarantees on the accuracy of the resulting graph.

\begin{definition}\label{def:eps-sparsifier}
A $ 1 \pm \vareps$ spectral sparsifier of $\Gg$ is a graph $\Hg \subseteq \Gg$ such that for all $\:x\in\Re^n$
\begin{align*}
(1-\vareps)\:x^\transp L_{\Gg} \:x \leq \:x^\transp L_\Hg \:x \leq (1 + \vareps) \:x^\transp L_{\Gg} \:x.
\end{align*}
\end{definition}

The key idea~\cite{spielman2011graph} is that to construct a sparse graph $\Hg$, it is sufficient to randomly select $m' = \bigotime(n \log(n)/\vareps^2)$ edges from $\Gg$ with a probability proportional to their effective resistance and add them to the new graph with suitable weights.
While storing the sparsified graph requires only $\bigotime(m')$ space, it still suffers from major limitations: \textit{(i)} the naive computation of the effective resistances needs $\bigotime(m n \log(n))$ time\footnote{A completely naive method would solve $n$ linear problems, each costing $\bigotime(mn)$. Using random projections we can solve only $\log(n)$ problems with only a small constant multiplicative error \cite{koutis2012improved}.}, \textit{(ii)} it requires $\bigotime(m)$ space to store the initial graph $\Gg$, and \textit{(iii)} computing the \hfs solution on $\Hg$ in a naive way still has a cost of $\bigotime(m'n)$. For this reason, we employ more sophisticated solutions and integrate the recent spectral sparsification technique for the semi-streaming setting in~\cite{kelner_spectral_2013} and the efficient solver for SDD systems in~\cite{koutis2011a-nearly-m}. The resulting algorithm is illustrated in Fig.~\ref{alg:sparse_ssl}.

\begin{figure}
\bookboxx{
\begin{algorithmic}
\begin{small}
\INPUT A sparsifier $\Hg$, block $\Delta$, number of edges $N$, effective resistance accuracy $\alpha$
    \OUTPUT A sparsifier $\Hg'$,  probabilities $\{\wt{p}'_e: e \in \Hg'\}$.
    \STATE Compute estimates of $\wt{R}'_e$ for any edge in $\Hg+\Delta$ such that $1/\alpha \leq \wt{R}'_e/R'_e \leq \alpha  $
    with an SDD solver \cite{koutis2011a-nearly-m}
    \STATE Compute probabilities $\wt{p}'_e = (a_e \wt{R}'_e)/(\alpha(n-1))$ and weights $w_e = a_e/(N\wt{p}'_e)$
    \STATE For all edges $e\in\Hg$ compute $\wt{p}'_e \leftarrow \min\{\wt{p}_e,\wt{p}'_e\}$ and initialize $\Hg' = \emptyset$
    \FOR{all edges $e \in \Hg$}
    \STATE Add edge $e$ to $\Hg'$ with weight $w_e$ with prob.~$\wt{p}'_e/\wt{p}_e$
    \ENDFOR
    \FOR{all edges $e \in \Delta$}
        \FOR{$ i = 1 $ to $N$}
        \STATE Add edge $e$ to $\Hg'$ with weight $w_e$ with prob.~$\wt{p}'_e$
        \ENDFOR
    \ENDFOR
    \end{small}
\end{algorithmic}
}
\vspace{-0.1in}
\caption{\small Kelner-Levin Sparsification algorithm~\cite{kelner_spectral_2013}\vspace{-0.1in}}\label{alg:kl_resparsify}
\end{figure}

We first introduce additional notation. Given two graphs $\Gg$ and $\Gg'$ over the same set of nodes $\X$, we denote by $\Gg+\Gg'$ the graph obtained by summing the weights on the edges of $\Gg'$ to $\Gg$. 
For any node $i=1,\ldots,n$, we denote with $\chi_i\in\Re^n$ the indicator vector so that $\chi_i - \chi_j$ is the ``edge'' vector. The effective resistance of an edge $e_{i,j}$ in a graph $\Gg$ is equal to $R_{e_{i,j}} = (\chi_i - \chi_j)^\transp L_\Gg^+ (\chi_i - \chi_j)$. The key intuition behind our \hfssparse is that processing  the graph incrementally allows to dramatically reduce the memory requirements and keep low time complexity at the same time. Let $\vareps$ be the (spectral) accuracy desired for the final sparsified graph $\Hg$, \hfssparse first partitions the set of edges $\edgeset$ of the original graph $\Gg$ into $\tau$ blocks $(\Delta_1,\ldots,\Delta_\tau)$ of size $N = \alpha^2 n\log^2(n) / \vareps^2$, with $\alpha = 1/(1-\vareps)$. While the original graph with $m$ edges is too large to fit in memory, each of these blocks has a number of edges which is nearly linear in the number of nodes and can be easily managed. \hfssparse processes blocks over iterations. Starting with an empty graph $\Hg_0$, at each iteration $t$ a new block~$\Delta_t$ is loaded and the intermediate sparsifier~$\Hg_{t-1}$ is updated using the routine \textsc{sparsify}, which is guaranteed to return a $(1\pm\vareps)$-sparsifier of size $N$ for the graph~$\Hg_{t-1}+\Delta_t$. After all the blocks are processed, a sparsifier $\Hg$ is returned and the \hfsstable solution can be computed. Since~$\Hg$ is very sparse (i.e., it only contains $N = \bigotime(n\log^2(n))$ edges), it is now possible to use efficient solvers for linear sparse systems and drastically reduce the computational complexity of solving \hfsstable from $\bigotime(Nn)$ down to $\bigotime(N\log(n))$. The routine \textsc{sparsify} can be implemented using different spectral sparsification techniques developed for the streaming setting, here we rely on the method proposed in~\cite{kelner_spectral_2013}. The effective resistance is computed for all edges in the current sparsifier and the new block using random projections and an efficient solver for SDD systems \cite{koutis2011a-nearly-m}. This step takes $\bigotime(N\log n)$ time and it returns $\alpha$-accurate estimates $\wt{R}'_e$ of the effective resistance for the $2N$ nodes in $\Hg_{t-1}$ and $\Delta_t$. If $\alpha = 1/(1-\vareps)$ and the input graph $\Hg_{t-1}$ is a $(1\pm\vareps)$-sparsifier, then sampling $N$ edges proportionally to $\wt{R}'_e$ is guaranteed to generate a $(1\pm\vareps)$-sparsifier for the full graph $\Gg_t = \sum_{s=1}^t \Delta_s$ up to iteration $t$. More details on this process are provided in Fig.~\ref{alg:kl_resparsify} and in~\cite{kelner_spectral_2013}. The resulting process has a space complexity $\bigotime(N)$ and a time complexity that never exceeds $\bigotime(N\log(n))$ in sparsifying each block and computing the final solution (see next section for more precise statements on time and space complexity).

\vspace{-0.05in}
\section{Theoretical Analysis}
\vspace{-0.05in}

We first report the time and space complexity of \hfssparse. This result follows from the properties of the sparsifier in~\cite{kelner_spectral_2013} and the SDD solver in~\cite{koutis2011a-nearly-m} and thus we do not report its proof.

\begin{lemma}\label{lem:sparse.complexity}
    Let $\vareps>0$ be the desired accuracy and $\delta>0$ the probability of failure. For any connected graph $\Gg=(\X,\edgeset)$ with $n$ nodes, $m$ edges, eigenvalues $0 = \lambda_1(\Gg) < \lambda_2(\Gg) \leq \ldots \leq \lambda_n(\Gg)$, and any partitioning of~$\edgeset$ into $\tau$ blocks, \hfssparse returns a graph $\Hg$ such that for any $i=1,\ldots,n$
\begin{align}\label{eq:spectrum.sparsifier}
(1-\vareps) \lambda_i(\Gg) \leq \lambda_i(\Hg) \leq (1+\vareps)\lambda_i(\Gg),
\end{align}
with prob.~$1-\delta$ (w.r.t.\@ the random estimation of the effective resistance and the sampling of edges in the \textsc{sparsify} routine). Furthermore, let $N = \alpha^2 n\log^2(n)/\vareps^2$ for $\alpha = 1/(1-\vareps)$ and $\tau = m / N$, then with prob.~$1-\delta$ \hfssparse has an amortized time per edge of $\bigotime(\log^3(n))$ and it requires $\bigotime(N)$ memory.\footnote{In all these big-$\bigotime$ expressions we hide multiplicative constants independent from the graph and terms $\log(1/\delta)$ which depends on the high-probability nature of the statements.}
\end{lemma}

The previous lemma shows the dramatic improvement of \hfssparse w.r.t.~\hfsstable in terms of both time and space complexity. In fact, while solving \hfsstable in a naive way can take up to $\bigotime(m)$ space and $\bigotime(mn)$ time, \hfssparse drops these requirements down to $\bigotime(n \log^2(n)/\vareps^2)$ space and $\bigotime(m \log^3(n))$ time, which allows scaling \hfsstable to graphs orders of magnitude bigger. These improvements have only a limited impact on the spectrum of~$\Gg$ and all its eigenvalues are approximated up to a $(1\pm\vareps)$ factor. Moreover, all of the sparsification guarantees hold w.h.p.\@ for any graph, regardless of how it is generated, its original spectra, and more importantly regardless of the exact order in which the edges are assigned to the blocks. Finally, we notice that the choice of the number of blocks as $m/N$ is crucial to guarantee a logarithmic amortized time, since each iteration takes $\bigotime(N\log^3(n))$ time. As discussed in Sect.~\ref{s:conclusions}, this property allows to directly apply \hfssparse in online learning settings where edges arrive in a stream and intermediate solutions have to be computed incrementally.
In the following, we show that, unlike other heuristics, the space complexity improvements obtained with sparsification come with guarantees and do not degrade the actual learning performance of \hfs. The analysis of SSL algorithms is built around the algorithm stability theory~\cite{bousquet_stability_2002}, which is extensively used to analyse transductive learning algorithms~\cite{el-yaniv_stable_2006,cortes2008stability}. 
We first remind the definition of algorithmic stability.

\begin{definition}\label{def:beta-stability}
    Let $\mathcal{L}$ be a transductive learning algorithm. We denote by $\:f$ and $\:f'$ the functions obtained by running $\mathcal{L}$ on datasets $\X = (\trainingset, \testset)$  and $\X = (\trainingset', \testset')$ respectively. $\mathcal{L}$ is uniformly $\beta$-stable w.r.t. the squared loss if there exists $\beta \geq 0$ such that for any two partitions $(\trainingset, \testset)$ and $(\trainingset', \testset')$ that differ by exactly one training (and test) point and for all $\:x \in \X$,
\begin{align*}
|(\:f(\:x) - \:y(\:x))^2 - (\:f'(\:x) - \:y(\:x))^2| \leq \beta.
\end{align*}
\end{definition}
\noindent Define the empirical error as $\wh{R}(\:f) = \tfrac{1}{l} \sum_{i=1}^{l} (\:f(x_i) - \:y(x_i))^2$ and the generalization error as $R(\:f) = \tfrac{1}{u}\sum_{i=1}^{u} (\:f(x_i) - \:y(x_i))^2$.
\begin{theorem}\label{thm:sparse-ssl-generalization}
    Let $\Gg$ be a fixed (connected) graph with $n$ nodes $\X$ and $m$ edges $\edgeset$ and eigenvalues $0 = \lambda_1(\Gg) < \lambda_2(\Gg) \leq \ldots \leq \lambda_n(\Gg)$. Let $\:y\in\Re^n$ be the labels of the nodes in $\Gg$ with $|\:y(x)| \leq M$ and $\funcspace$ be the set of centered functions such that $|\:f(x) - \:y(x)| \leq c$. Let $\calS\subset \X$ be a random subset of labeled nodes. If the corresponding labels $\wt{\:y}_S$ are centered and \hfssparse is run with parameter $\vareps$, then w.p.~at least $1 - \delta$ (w.r.t. the random generation of the sparsifier $\Hg$ and the random subset of labeled points $\calS$) the resulting function~$\wt{\:f}$ satisfies,
\begin{align}\label{eq:bound}
    R(\wt{\:f}) \leq \widehat{R}(\wh{\:f}) + &\frac{ l^2 \gamma^2 \lambda_n(\Gg)^2 M^2 \varepsilon^2 }{(l \gamma (1 - \varepsilon)\lambda_{2}(\Gg) - 1)^4}+ \beta +\nonumber\\
    &\left(2\beta + \frac{c^2(l+u)}{lu}\right)\sqrt{\frac{\pi(l,u)\ln\frac{1}{\delta}}{2}},
\end{align}
where $\wh{\:f}$ is the solution of exact \hfsstable on $\Gg$,
\begin{align*}
    &\pi(l,u) = \frac{lu}{l+u-0.5} \frac{2\max\{l,u\}}{2\max\{l,u\}-1}, \enspace \text{ and }\\
    &\beta \leq \frac{1.5 M\sqrt{l}}{(l \gamma (1-\varepsilon)\lambda_{2}(\Gg)-1)^{2}} + \frac{4M}{l \gamma (1 - \varepsilon)\lambda_{2}(\Gg)-1}.
\end{align*}
\end{theorem}
Theorem~\ref{thm:sparse-ssl-generalization} shows how approximating $\Gg$ with $\Hg$ impacts the generalization error as the number of labeled samples $l$ increases. If we compare the bound to the exact case ($\varepsilon = 0$), we see that for any fixed $\varepsilon$ the rate of convergence is not affected by the sparsification. The first term in Eq.~\ref{eq:bound} is of order $\bigotime(\varepsilon^2/l^2(1-\varepsilon)^4)$ and it is the additive error w.r.t.\ the empirical error $R(\wh{\:f})$ of the \hfsstable solution. For any constant value of~$\varepsilon$, this term scales as $1/l^2$ and thus it is dominated by the second term in the stability $\beta$. The $\beta$ term itself preserves the same order of convergence as for the exact case up to a constant term of order $1/(1-\varepsilon)$. In conclusion, for any fixed value of the $\varepsilon$, \hfssparse preserves the same convergence rate as the exact \hfsstable w.r.t.\@ the number of labeled and unlabeled points. This means that $\vareps$ can be arbitrarily chosen to trade off accuracy and space complexity (in Lemma~\ref{lem:sparse.complexity}) depending on the problem constraints. Furthermore running time does not depend on this trade-off, because less frequent resparsifications will balance the increased block size.

\begin{proof}

\textbf{Step 1 (generalization of stable algorithms).}
Let $\beta$ be the stability of \hfssparse, then using the result in~\cite{cortes2008stability}, we have that with probability at least $1 - \delta$ (w.r.t.\@ the randomness of the labeled set $\trainingset$) the solution $\wt{\:f}$ returned by the \hfssparse satisfies
\begin{align*}
R(\wt{\:f}) \leq \widehat{R}(\wt{\:f}) + \beta + \Big(2\beta + \frac{c^2(l+u)}{lu}\Big)\sqrt{\frac{\pi(l,u)\log(1/\delta)}{2}}.
\end{align*}
In order to obtain the final result and study how much the sparsification may affect the performance of \hfsstable, we first derive an upper bound on the stability of \hfssparse and then relate its empirical error to the one of \hfsstable.

\textbf{Step 2 (stability).}
The bound on the stability follows similar steps as in the analysis of \hfsstable in~\cite{belkin2004regularization} integrated with the properties of streaming spectral sparsifiers in~\cite{kelner_spectral_2013} reported in Lemma~\ref{lem:sparse.complexity}.
Let $\trainingset$ and $\trainingset'$ be two labeled sets only differing by one element and $\wt{\:f}$ and $\wt{\:f}'$ be the solutions obtained by running \hfssparse on $\calS$ and $\calS'$ respectively. Without loss of generality, we assume that $I_{\trainingset}(l,l) = 1$ and $I_{\trainingset}(l+1,l+1) = 0$, and the opposite for $I_{\trainingset'}$. The original proof in~\cite{cortes2008stability} showed that the stability $\beta$ can be bounded as $\beta\leq\normsmall{\wt{\:f} - \wt{\:f}'}$. In the following we show that the difference between the solutions $\wt{\:f}$ and $\wt{\:f}'$, and thus the stability of the algorithm, is~strictly related to eigenvalues of the sparse graph$\Hg$. Let $A=P_{\funcspace}(l \gamma L_{\Hg}+I_{\trainingset})$ and $B=P_{\funcspace}(l \gamma L_{\Hg}+I_{\trainingset'})$, we remind that if the labels are centered, the solutions of \hfssparse can be conveniently written as $\wt{\:f}=A^{-1}\wt{\:y}_{\trainingset}$ and $\wt{\:f}'=B^{-1}\wt{\:y}_{\trainingset'}$. As a result, the difference between the solutions can be written as
\begin{align}\label{eq:stability.step1}
    \Vert\wt{\:f}-\wt{\:f}'\Vert&=\Vert A^{-1}\wt{\:y}_{\trainingset}-B^{-1}\wt{\:y}_{\trainingset'}\Vert\\
    &\leq \Vert A^{-1}(\wt{\:y}_{\trainingset}-\wt{\:y}_{\trainingset'})\Vert+\Vert A^{-1}\wt{\:y}_{\trainingset'}-B^{-1}\wt{\:y}_{\trainingset'}\Vert.\nonumber
\end{align}
Let consider any vector $\:f \in \funcspace$, since the null space of a Laplacian $L_{\Hg}$ is the one vector $\:1$ and $P_\funcspace = L_{\Hg}L_{\Hg}^+$, then $P_\funcspace \:f = \:f$. Thus we have
\begin{align}
    &\Vert P_{\funcspace}(l\gamma L_{\Hg}+I_{\calS})\:f\Vert\stackrel{(1)}{\geq}\Vert P_{\funcspace}l\gamma L_{\Hg}\:f\Vert \!-\! \Vert P_{\funcspace}I_{\calS}\:f\Vert\nonumber\\
    &\stackrel{(2)}{\geq} \Vert P_{\funcspace}l\gamma L_{\Hg}\:f\Vert \!-\! \Vert\:f\Vert \stackrel{(3)}{\geq}(l \gamma \lambda_{1}(\Hg)\!-\!1)\Vert \:f\Vert \label{eq:lower.bound}
\end{align}
where $(1)$ follows from the triangle inequality 
and $(2)$ follows from the fact that $\Vert P_{\funcspace}I_{\calS}\:f\Vert \leq \Vert\:f\Vert$ since the largest eigenvalue of the project matrix $P_\funcspace$ is one and the norm of $\:f$ restricted on $\calS$ is smaller than the norm of $\:f$. Finally (3) follows from the fact that $\Vert P_\funcspace L_\Hg \:f \Vert = \Vert L_\Hg L_\Hg^+ L_\Hg \:f \Vert = \Vert L_\Hg \:f\Vert$ and since $\:f$ is orthogonal to the null space of $L_\Hg$ then $\Vert L_\Hg \:f\Vert \geq \lambda_2(\Hg) \Vert\:f\Vert$, where $\lambda_2(\Hg)$ is the smallest non-zero eigenvalue of $L_\Hg$. At this point we can exploit the spectral guarantees of the sparsified graph $L_\Hg$ and from Lemma~\ref{lem:sparse.complexity}, we have that $\lambda_2(\Hg) \geq (1-\varepsilon)\lambda_2(\Gg)$. 
As a result, we have an upper-bound on the spectral radius of the inverse operator $(P_{\funcspace}(l\gamma L_{\Hg}+I_{\trainingset}))^{-1}$ and thus
\begin{align*}
    \Vert A^{-1}(\:y_{\trainingset}-\:y_{\trainingset'})\Vert&\leq \frac{\Vert\:y_{\trainingset}-\:y_{\trainingset'}\Vert}{l\gamma (1 - \varepsilon)\lambda_{1}(\Gg)-1} \\
    &\leq \frac{4M}{l\gamma (1 - \varepsilon)\lambda_{1}(\Gg)-1},
\end{align*}
where the first step follows from Eq.~\ref{eq:lower.bound} since both $\wt{\:y}_{\trainingset}$ and $\wt{\:y}_{\trainingset'}$ are centered and thus $(\:y_{\trainingset}-\:y_{\trainingset'})\in\funcspace$, and the second step is obtained by bounding $\Vert\wt{\:y}_{\trainingset}-\wt{\:y}_{\trainingset'}\Vert\leq \Vert\:y_{\trainingset}-\:y_{\trainingset'}\Vert + \Vert\ol{\:y}_{\trainingset}-\ol{\:y}_{\trainingset'}\Vert \leq 4M$. The second term in Eq.~\ref{eq:stability.step1} can be bounded as
\begin{align*}
    &\Vert A^{-1}\wt{\:y}_{\trainingset'}-B^{-1}\wt{\:y}_{\trainingset'}\Vert=\Vert B^{-1}(B-A)A^{-1}\wt{\:y}_{\trainingset'}\Vert\\
&=\Vert B^{-1}P_{\funcspace}(I_{\trainingset}-I_{\trainingset'})A^{-1}\wt{\:y}_{\trainingset'}\Vert\leq\frac{1.5 M\sqrt{l}}{(l \gamma (1-\varepsilon)\lambda_{1}(\Gg)-1)^{2}},
\end{align*}
where we used $\Vert\wt{\:y}_{\trainingset'}\Vert\leq \Vert\:y_{\trainingset'}\Vert + \Vert\ol{\:y}_{\trainingset'}\Vert \leq 2M\sqrt{l}$, $\Vert P_{\funcspace}(I_{\trainingset}-I_{\trainingset'})\Vert \leq \sqrt{2} < 1.5$ and we applied Eq.~\ref{eq:lower.bound} twice.
Putting it all together we obtain the final bound reported in the statement.

\textbf{Step 3 (empirical error).} The other element effected by the sparsification is the empirical error $\wh{R}(\wt{\:f})$. Let $\wt{A} = P_{\funcspace}(l \gamma L_{\Hg}+I_{\trainingset})$, $\wh{A} = P_{\funcspace}(l \gamma L_{\Gg}+I_{\trainingset})$, then 
\begin{align*}
    \wh{R}(\wt{\:f}) &= \frac{1}{l} \normsmall{I_\trainingset \wt{\:f} -I_\trainingset \wh{\:f} +I_\trainingset \wh{\:f}  - \wt{\:y}_{\trainingset}}^2\nonumber\\
&\leq \tfrac{1}{l} \normsmall{I_\trainingset \wh{\:f}  - \wt{\:y}_{\trainingset}}^2 + \tfrac{1}{l} \normsmall{I_\trainingset \wt{\:f} -I_\trainingset \wh{\:f} }^2 \\
&\leq \wh{R}(\wh{\:f}) + \tfrac{1}{l} \normsmall{I_\trainingset(\wt{A}^{-1} -\wh{A}^{-1}) \wt{\:y}_{\trainingset} }^2\nonumber\\
&\leq \wh{R}(\wh{\:f}) + \tfrac{1}{l} \normsmall{\wh{A}^{-1}(\wh{A} -\wt{A})\wt{A}^{-1} \wt{\:y}_{\trainingset} }^2\\
&\leq \wh{R}(\wh{\:f}) + \frac{1}{l}\frac{lM^2}{(l \gamma (1 - \varepsilon)\lambda_{1}(\Gg) - 1)^4} \normsmall{\wh{A} -\wt{A}}^2,
\end{align*}
where in the last step we applied Eq.~\ref{eq:lower.bound} on both $\wh{A}^{-1}$ and $\wt{A}^{-1}$. We are left with $\normsmall{\wh{A} -\wt{A}}^2 = \normsmall{P_{\funcspace} l \gamma (L_{\Gg} - L_{\Hg})}^2$. We first recall that $P_\funcspace = L_\Gg^+ L_\Gg = L_{\Gg}^{-1/2}L_{\Gg}L_{\Gg}^{-1/2}$ (and equivalently with $\Gg$ replaced by $\Hg$) and we introduce $\wt{P}_{\funcspace} = L_{\Gg}^{-1/2}L_{\Hg}L_{\Gg}^{-1/2}$. We have
\begin{align*}
    \normsmall{\wh{A} -\wt{A}}^2 &\stackrel{(1)}{=} l^2 \gamma^2 \normsmall{ L_{\Gg} - L_{\Hg}}^2\\
    &\stackrel{(2)}{=} l^2 \gamma^2 \normsmall{L_{\Gg}^{1/2}(P_{\funcspace} - \wt{P}_{\funcspace})L_{\Gg}^{1/2}}^2 \\
&\stackrel{(3)}{\leq} l^2 \gamma^2 \lambda_n(\Gg)^2 \normsmall{P_{\funcspace} - \wt{P}_{\funcspace}}^2 \stackrel{(4)}{\leq} l^2 \gamma^2 \lambda_n^2 \varepsilon^2,
\end{align*}
where in $(1)$ we use $P_{\funcspace} L_\Gg = L_\Gg$ and $P_{\funcspace} L_\Hg = L_\Hg$, in $(2)$ we rewrite $L_\Gg = L_\Gg^{1/2}L_\Gg^{-1/2}L_\Gg L_\Gg^{-1/2}L_\Gg^{1/2} = L_\Gg^{1/2} P_\funcspace L_\Gg^{1/2}$ and $L_\Gg = L_\Gg^{1/2}L_\Gg^{-1/2}L_\Hg L_\Gg^{-1/2}L_\Gg^{1/2} = L_\Gg^{1/2} \wt{P}_\funcspace L_\Gg^{1/2}$, in $(3)$ we split the norm and use the fact that the spectral norm of $L_\Gg$ corresponds to its largest eigenvalue $\lambda_n(\Gg)$, while in $(4)$ we use the fact that Def.~\ref{def:eps-sparsifier} implies that $(1-\vareps) P_{\funcspace} \preceq \wt{P}_{\funcspace} \preceq (1 + \vareps) P_{\funcspace}$ and thus the largest eigenvalue of $P_{\funcspace} - \wt{P}_{\funcspace}$ is $\varepsilon^2 \Vert P_\funcspace\Vert \leq \vareps^2$.
The final statement follows by combining the three steps above.
\end{proof}

\vspace{-0.1in}
\section{Experiments}\label{s:experiments}

In this section we evaluate the empirical accuracy of \hfssparse compared to other baselines for large-scale SSL on both synthetic and real datasets.

\begin{figure*}[t]
    \vspace{-0.3cm}
    \centering
    \begin{tabular}{c|c|c|c|c|}
    & Guarant. & Space & Preprocessing Time & Solving Time\\
    \hline
    \vspace{-0.3cm}& & & & \\
    \hfssparse & \includegraphics[height=0.3cm]{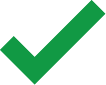} & $N = \bigotime(n \log^2(n)) $& $\bigotime(m \log^3(n)) $& $\bigotime(N \log(n)) = \bigotime(n \log^3(n))$\\
    \hfsstable & \includegraphics[height=0.3cm]{guarantees_yes}& \rcolbm{\bigotime(m)}& $\bigotime(m) $& $\bigotime(m n)$\\
    \simplehfs &\includegraphics[height=0.3cm]{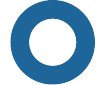} & \rcolbm{\bigotime(m)}& $\bigotime(mq) $ & $\bigotime(q^4)$\\
    \fergus & \includegraphics[height=0.3cm]{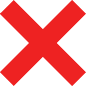}& $\bigotime(nd + nq + b^2) $& $\bigotime(qb^3  + db^3) $& $\bigotime(q^3 + nq)$\\
    \subs & \includegraphics[height=0.3cm]{guarantees_no}& $\bigotime(sk) $& $\bigotime(m) $& $\bigotime(s^2 k + n)$
    \end{tabular}
    \caption{{\small \textbf{Guarantees and Computational complexities.}
    Bold text indicates unfeasible time or space complexity.
    \protect \includegraphics[height=0.2cm]{guarantees_no} Guarantees unavailable.
     \protect \includegraphics[height=0.2cm]{guarantees_kinda} \simplehfs's guarantees require assumptions on the graph $\Gg$. \vspace{-0.3cm}}\label{fig:comparison-table}}
\end{figure*}

\begin{figure}[t]
\begin{tabular}{m{0.3\columnwidth}m{0.7\columnwidth}}
        \sidesubfloat[]{\includegraphics[height=6cm]{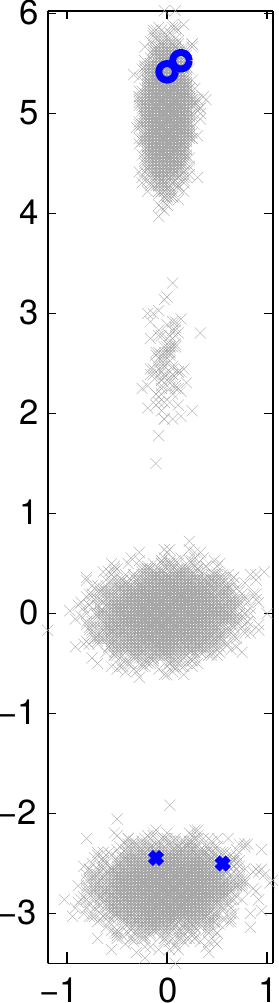} \label{fig:gen-data}}&
        \sidesubfloat[]{\includegraphics[height=3cm]{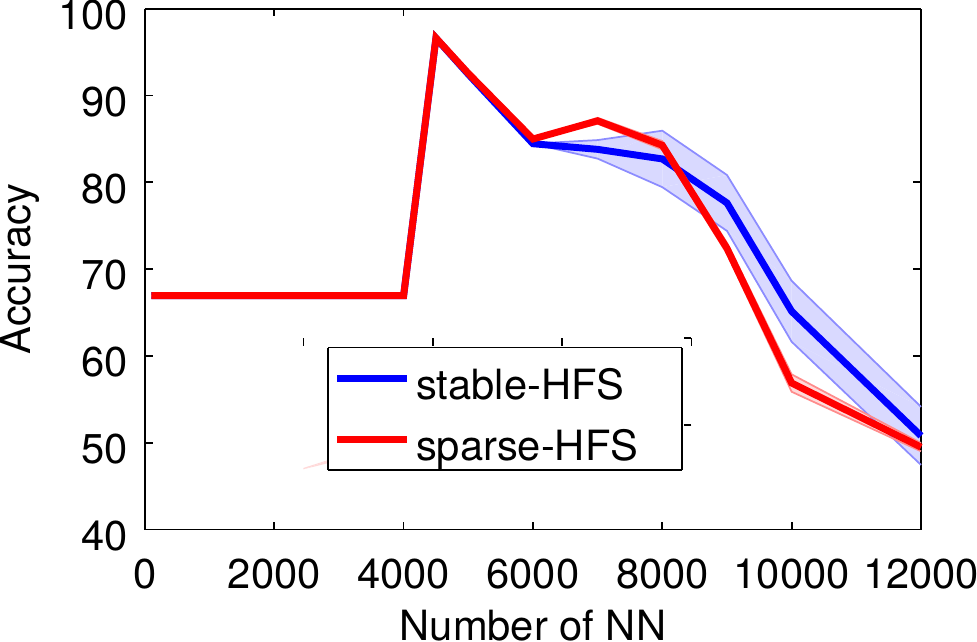}\label{fig:gen-error}}\newline
        \sidesubfloat[]{\includegraphics[height=3cm]{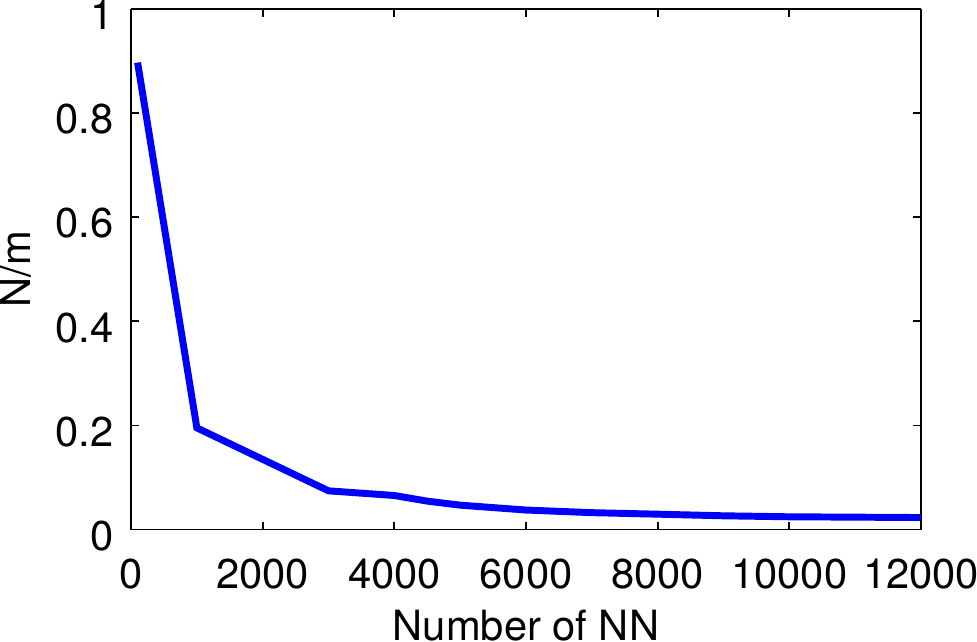}\label{fig:edge-ratio}}
    \end{tabular}
\caption{\small \protect\subref{fig:gen-data} The dataset of the synthetic experiment, \protect\subref{fig:gen-error} Accuracy of \hfsstable and \hfssparse, \protect\subref{fig:edge-ratio} ratio of the number of edges $|\Hg|/|\Gg|$.\vspace{-0.25in}}
\label{fig:fibsyn}
\end{figure}

\textbf{Synthetic data.}
The objective of this first experiment is to show that the sparsification method is effective in reducing the number of edges in the graph and that preserving the full spectrum of $\Gg$ retains the accuracy of the exact \hfs solution.
We evaluate the algorithms on the $\Re^2$ data distributed as in
Fig.~\ref{fig:fibsyn}\subref{fig:gen-data}, which is designed so that a large
number of neighbours is needed to achieve a good accuracy. The dataset is
composed of $n = 12100$ points, where the two upper clusters belong to one
class and the two lower to the other. We build an unweighted, $k$-nn graph
$\Gg$ for $k = {100,\ldots,12000}$.
After constructing the graph, we randomly select two points from the uppermost
and two from lowermost cluster as our labeled set $\trainingset$. We then run
\hfssparse with $\vareps = 0.8$ to compute $\Hg$ and~$\wt{\:f}$, and run
(exact) \hfsstable
on $\Gg$ to compute~$\wh{\:f}$, both with $\gamma = 1$.
Fig.~\ref{fig:fibsyn}\subref{fig:gen-error} reports the accuracy of the two
algorithms. Both algorithms fail to recover a good solution until
$k \approx 4000$. This is due to the fact that until a certain threshold, each cluster remains separated and the labels
cannot propagate. Beyond this threshold, \hfsstable is very
accurate, while, as $k$ increases again, the graph becomes almost full, masking
the actual structure of the data and thus loosing performance again. We notice
that the accuracy of \hfsstable and \hfssparse is never significantly different,
and, quite importantly, they match around the value of $k = 4500$
that provides the best performance. This is
in line with the theoretical analysis that shows that the contribution due to
the sparsification error has the same order of magnitude as the other elements
in the bound.
Furthermore, in Fig.~\ref{fig:fibsyn}\subref{fig:edge-ratio} we report the ratio of the number of edges in the sparsifier~$\Hg$ and $\Gg$. 
This quantity is always smaller than one and it constantly decreases since the number of edges in $\Hg$ is constant, while the size $\Gg$ increases linearly with the number of neighbors (i.e., $|\Hg|/|\Gg| = \bigotime(1/k)$).
We notice that for the optimal $k$ the sparsifier contains less than 10\% of the edges of the original graph but it achieves almost the same accuracy.

\begin{figure}[t]
\begin{center}
    \includegraphics[width=\columnwidth]{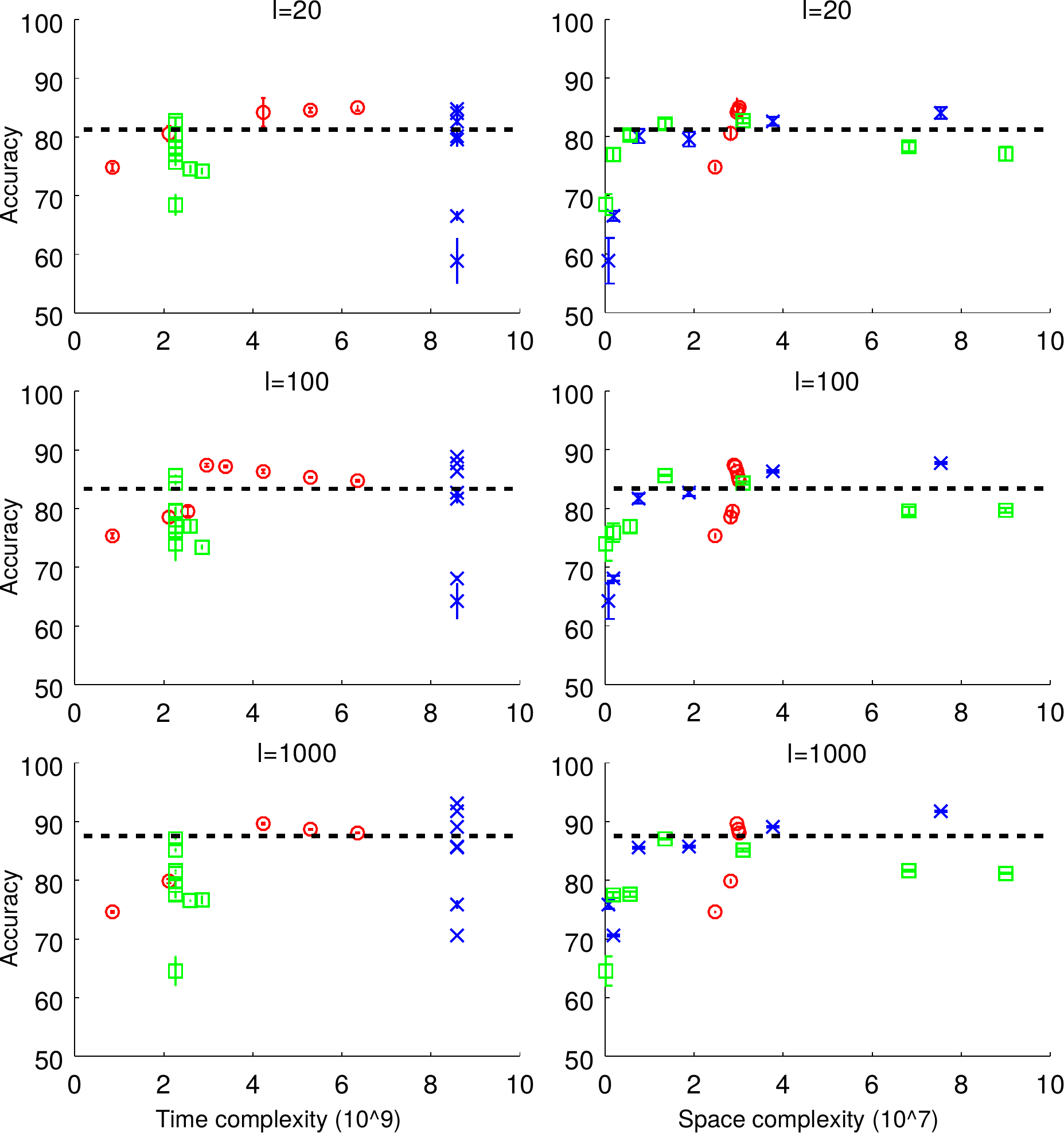}
\end{center}
\vspace{-0.05in}
\caption{\small Accuracy vs complexity on the TREC 2007 SPAM Corpus for different number of labels. \textit{Legend:} \allowbreak \protect\includegraphics[height=0.2cm]{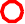} \hfssparse, \protect\includegraphics[height=0.2cm]{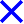} \fergus, \protect\includegraphics[height=0.2cm]{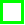} \subs, \protect\raisebox{0.2em}{\protect\includegraphics[width=0.5cm]{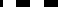}} \onenn \vspace{-0.5cm}\label{fig:gen-error-emp}}
\end{figure}

\textbf{Spam-filtering dataset.} We now evaluate the performance of our
algorithm on the TREC 2007 Public Spam Corpus\footnote{\scriptsize\url{http://plg.uwaterloo.ca/~gvcormac/treccorpus07/}},
that contains $n = 75419$ raw emails labeled as either \textit{SPAM} or
\textit{HAM}. The emails are provided as raw text and we applied standard NLP
techniques to extract features vectors from it. In particular, we computed
TF-IDF scores for each of the emails, with some additional cleaning in the
form of a stop word list, simple stemming and dropping the 1\% most common and
most rare words. We ended up with~$d = 68697$ features, each representing a
word present in some of the emails. From these features we proceeded to build
a graph $\Gg$ where given two emails $\:x_i,\:x_j$, the weight is computed as
$a_{ij} = \exp(-\Vert \:x_i - \:x_j\Vert /2\sigma^2)$, with $\sigma^2=3$. We
consider the transductive setting, where the graph is fixed and known, but
only a small random subset of $l = \{20,100,1000\}$ labels is revealed to the
algorithm. As a performance measure, we consider the prediction accuracy over
the whole dataset. We compare our method to several baselines. The most basic
supervised baseline is \onenn, which connects each node to the closest labeled node. The
\subs algorithm selects uniformly $s$ nodes out of~$n$, computes the \hfs
solution on the induced subgraph of $\Gg$ and assigns to each node outside of the subset the same label as the closest node in the subset.
\subs's complexity depends on the size
$s$ of the subgraph and the number $k$ of neighbors retained when building the~$k$-nn subgraph. The eigenfunction (\fergus)
algorithm~\cite{fergus2009semi-supervised} tries to sidestep the
computational complexity of finding an \hfs solution on $\Gg$,
by directly approximating the distribution that created the graph.
Starting directly from the samples, each of the $d$ feature's density is separately approximated
using histograms with $b$ bins.
From the histograms, $q$ empirical eigenfunctions (vectors in $\Real^n$) are extracted and used
to compute the final solution.
We did not include \hfsstable and \simplehfs in the comparison
because their $\bigotime(m)$ space complexity made them unfeasible for this dataset.
In Fig.~\ref{fig:gen-error-emp}, we
report the accuracy of each method against the time and space complexity,
where each separate point corresponds to a different choice in metaparameters (e.g.~$k,q,s$).
For \fergus, we use the same $b = 50$ as in the original implementation, but
we varied $q$ from 10 to 2000. For \subs, $s=15000$ and $k$ varies from 100 to
10000. We run \hfssparse on $\Gg$ setting $\varepsilon = 0.9$, and
changing the size $m$ of the input graph by changing the number of
neighbours $k$ from 1000 to 7500.
Since the actual running time and memory
occupation are highly dependent on the implementation (e.g., \fergus is
implemented in Matlab, while \hfssparse is Matlab/C), the
complexities are computed using their theoretical form (e.g.,
$\bigotime(m \log^3(n))$ for \hfssparse)
with the values actually used in the experiment (e.g.,~$m = nk$ for a $k$-nn graph).
All the complexities are reported in Fig.~\ref{fig:comparison-table}.
The only exception is the number of edges in the sparsifier $N$ used in the space complexity of \hfssparse.
Since this is a random quantity that holds only w.h.p.~and that is
independent from implementation details, we measured it empirically and
used it for the complexities.
For all methods we notice that the performance increases as the space
complexity gets larger, until a peak is reached, while additional space
induces the algorithms to overfit
and reduces accuracy.
For \fergus this means that a large number of eigenfunctions is necessary
to accurately model the high dimensional distribution. And as theory predicts,
\subs's uniform sampling is not efficient to approximate the graph spectra,
and a large subset of the nodes is required for good performance.
\hfssparse's accuracy also increases as the input graph gets richer, but unlike
the other methods the space complexity does not change much. This is
because the sparsifier is oblivious to the structure of the graph,
and even if \hfssparse reaches its optimum performance for $k=3000$,
the sparsifier contains roughly the same number of edges present as
$k=1000$, and only 5\% of the edges present in the input graph.
Although preliminary, this
experiment shows that the theoretical properties of \hfssparse translate into
an effective practical algorithm which is competitive with state-of-the-art
methods for large-scale SSL.

\vspace{-0.15in}
\section{Conclusions and Future Work}\label{s:conclusions}
\vspace{-0.15in}

We introduced \hfssparse, an algorithm that combines
sparsification methods and efficient solvers for SDD
systems to find approximate \hfs solutions
using only $\bigotime(n\log^2(n))$ space instead of $\bigotime(m)$.
Furthermore, we show that the $\bigotime(m\log^3(n))$ time complexity of the methods
is only a $\polylog$ term away from the smallest possible complexity  $\Omega(m)$.
Finally, we provide a bound on the generalization error that
shows that the sparsification does not affect the asymptotic convergence rate
of \hfs. As such, the accuracy parameter $\vareps$ can be freely chosen to
meet the desired trade-off between accuracy and space complexity.
In this paper we relied on the sparsifier in~\cite{kelner_spectral_2013}
to guarantee a fixed space requirement, and
the solver in~\cite{koutis2011a-nearly-m} to efficiently compute
the effective resistances. Both are straightforward to scale and parallelize \cite{blelloch2010hierarchical},
and the bottleneck in practice reduces to finding a fast sparse matrix-vector
multiplication implementation
for which many off-the-shelf solutions exist. We also remark that \hfssparse could easily
accommodate any improved version of these algorithms and their properties
would directly translate into the performance of \hfssparse.
In particular \cite{koutis2011a-nearly-m} already mentions how finding
an appropriate spanning tree (a low-stretch tree) and using it
as the backbone of the sparsifier allows to reduce the space
requirements of the sparsifier.
Although this technique could lower the space complexity to
$\bigotime(n \log(n))$, it is not clear how to find such a tree incrementally.
An interesting feature of \hfssparse is that it could be easily employed in
online learning problems where edges arrive in a stream and intermediate
solutions have to be computed over time.
Since \hfssparse has a
$\bigotime(\log^3(n))$ amortized time per edge, it could compute intermediate
solutions every $N$ edges
without compromising its overall time complexity.
The fully dynamic setting, where edges can be both inserted and removed,
is an important extension where our approach could be further investigated,
especially because it has been observed in several domains that graphs
become denser as they evolve over time \cite{leskovec2005graphs}.
While sparsifiers have been developed for this setting (see
e.g.,~\cite{kapralov_single_2014}), current solutions would require
$\bigotime(n^2 \polylog(n))$ time to compute the \hfs solution, thus making it
unfeasible to repeat this computation many times over the stream. Extending
sparsification techniques to the fully dynamic setting in a computationally
efficient manner is  an open problem.

\bibliographystyle{plain}

\end{document}